\documentclass[10pt,twocolumn,letterpaper]{article}

\usepackage{cvpr}
\usepackage{times}
\usepackage{epsfig}
\usepackage{graphicx}
\usepackage{amsmath}
\usepackage{amssymb}

\usepackage[breaklinks=true,bookmarks=false]{hyperref}

\cvprfinalcopy 

\def\cvprPaperID{8643} 

\begin{document}

\title{Simultaneous Mapping and Target Driven Navigation}

\author{Georgios Georgakis, Yimeng Li, and  Jana Ko{\v{s}}eck{\'a}\\
Department of Computer Science, George Mason University, Fairfax VA\\
{\tt\small \{ggeorgak, yli44, kosecka\}@gmu.edu}
}


\maketitle

\begin{abstract}
This work presents a modular architecture for simultaneous mapping and target driven navigation in indoors environments. The semantic and appearance stored in 2.5D map is distilled from RGB images, semantic segmentation and outputs of object detectors by convolutional neural networks. Given this representation, the mapping module learns to localize the agent and register consecutive observations in the map. The navigation task is then formulated as a problem of learning a policy for reaching semantic targets using current observations and the up-to-date map. We demonstrate that the use of semantic information improves localization accuracy and the ability of storing spatial semantic map aids the target driven navigation policy. The two modules are evaluated separately and jointly on Active Vision Dataset~\cite{active-vision-dataset2017} and Matterport3D environments~\cite{Matterport3D}, demonstrating improved performance on both localization and navigation tasks.

\end{abstract}

\section{Introduction}
Navigation is one of the fundamental capabilities of autonomous agents. In recent years there was a surge of novel approaches, which study the problem of goal or target driven navigation using end-to-end learning strategies~\cite{zhu2017target,mirowski2016learning,das2018embodied,fang2019scene}.
These approaches use data-driven techniques for learning navigation policies deployable in previously unseen environments without constructing an explicit spatial representation of the environment. These models exploit the power of recurrent neural networks for learning predictions from sequences of observations.

The approaches that address the navigation and planning by learning spatial representations of the environment often assume perfect localization both in the training and testing stage~\cite{gupta2017cognitive,parisotto2017neural}.
The problem of localization and mapping is challenging on its own and existing approaches for learning spatial representations which are optimized for localization tasks ~\cite{henriques2018mapnet} have been shown to outperform traditional simultaneous localization and mapping methods (SLAM) on the localization task~\cite{SLAM16}. \\





\noindent
The presented work investigates the problem of simultaneous mapping and target driven navigation in previously unseen environments. 
The problem of target driven navigation is a problem of an agent finding its way through a complex environment to a target (e.g. go to the couch). The goal of our work is to exploit mapping and localization module to guide navigation strategies and relax the assumption of perfect localization and at the same time endow the map representation with richer semantic information. 
Towards this end we propose to build and use spatial  allocentric 2.5D map, which will facilitate both localization and semantic target navigation.
\begin{figure}[!t]
\begin{center}
\includegraphics[width=1\linewidth]{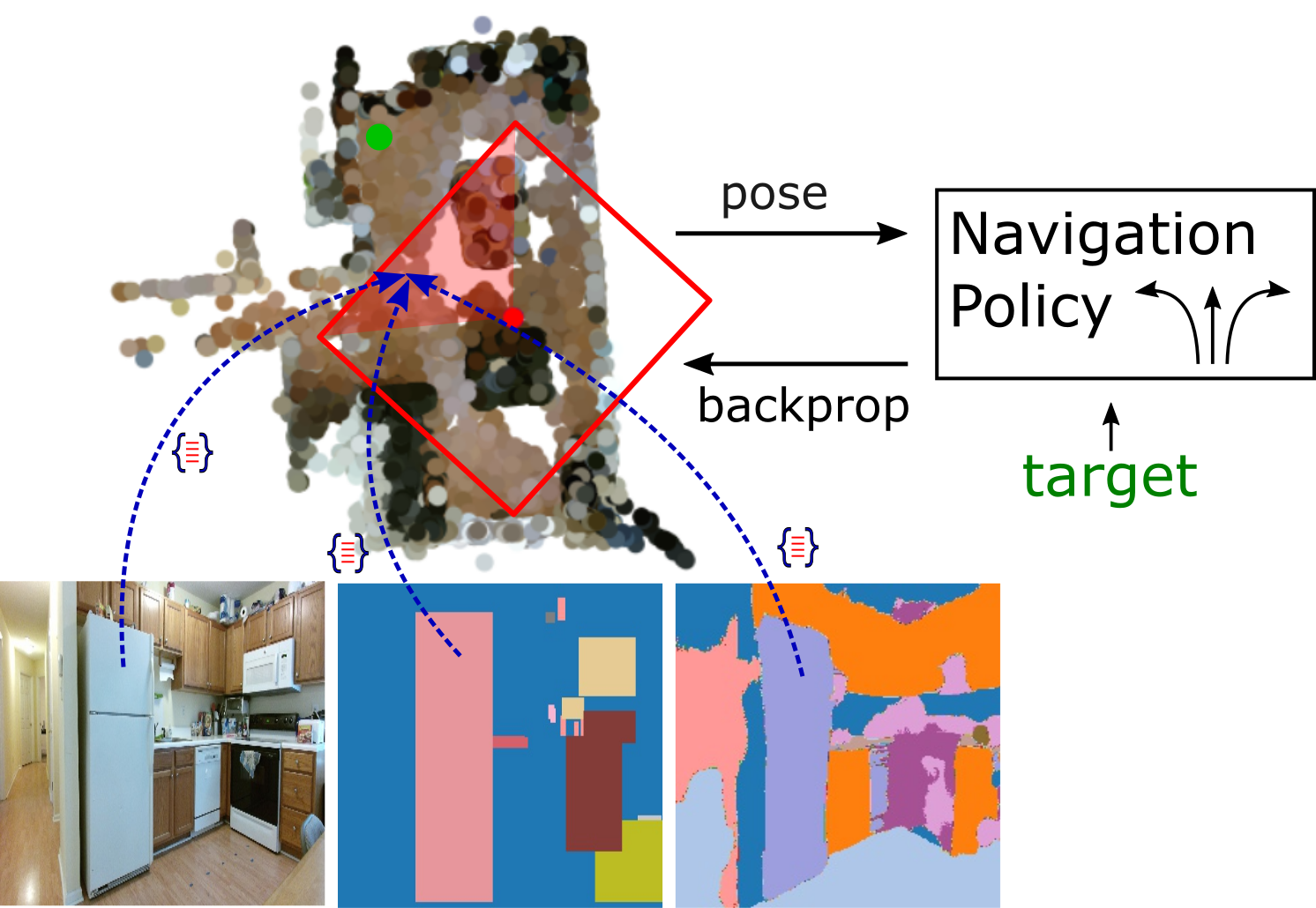}
\end{center}
   \caption{We present a new method for target driven navigation that leverages an 2.5D allocentric map with learned semantic representations suitable for both localization and semantic target driven navigation.}
\label{fig:title}
\end{figure}
Instead of using map representation derived directly from pixel values, we propose to learn suitable task related embeddings from outputs of an object detector and semantic segmentation. 

The proposed method consists of two modules. First, a semantic map inspired by MapNet architecture~\cite{henriques2018mapnet} is responsible for continuous localization and registration of agent's observations in the map. Second, is a navigation module that uses the partial map, predicted pose of the agent along with current observations for learning a
target reaching policy. 
In summary our contributions are as follows:
\begin{itemize}
    \item We extend the MapNet~\cite{henriques2018mapnet} approach and endow 2.5D memory with semantically informed features and demonstrate its improvement on the localization task.
    \item We show the effectiveness of spatial  mapping for target driven navigation and learn the navigation policies for the task. 
    \item We evaluate both localization and target driven navigation tasks on real Active Vision Dataset~\cite{active-vision-dataset2017} and Matterport3D~\cite{Matterport3D} environment, demonstrating superior performance compared to previous methods.
\end{itemize}


\section{Related Work}
Traditional approaches for mapping and navigation problem focus on 3D metric and semantic mapping of the environment~\cite{SLAM16} followed by path planning and control. They require building a 2D or 3D map ahead of time before planning, and do not exploit general semantics and contextual cues in the planning and decision stage.
In recent years there was a surge of novel approaches, which study the problem of goal or target driven navigation using end-to-end deep reinforcement learning and vary in the proposed architectures and reward structure to train the models~\cite{zhu2017target,mirowski2016learning,das2018embodied,fang2019scene, House3D_NIPS17} These methods use variations of Recurrent Neural networks (i.e. LSTM's) with the memory  implicitly represented by the hidden state of the model. Majority of the above mentioned methods do not have explicit notion of the map or spatial representation of the environment. 



The methods which explicitly learn a spatial representation of the environment, proposed task-dependent differentiable spatial memories to represent the environment~\cite{gupta2017cognitive, henriques2018mapnet, devendra2019neural, gordon2018iqa, parisotto2017neural, zhang2017neural}
and typically focus on goal or target driven navigation, localization or exploration tasks. For example, Henriques \etal~\cite{henriques2018mapnet} proposed an architecture that dynamically updates an agent's allocentric representation for the task of localization. In Gupta \etal~\cite{gupta2017cognitive} a mapping module fused information from learned image embeddings across multiple views in an egocentric top-view map of the environment. The mapping module 
was trained for goal point and semantic target based navigation tasks and assumed accurate localization both in training and testing stage. Authors in~\cite{devendra2019neural} train a policy that takes as input a predicted egocentric map and outputs long-term goal for a planner, while Gordon \etal~\cite{gordon2018iqa} uses a GRU to perform egocentric updates to a local window within a spatial memory given the agent's current location and viewpoint. The spatial memory contains object confidences at each location of a 2D grid, but it does not encode the 3D spatial capacity of the objects in the environment and thus cannot take advantage of multi-view information to deal with occlusions. The work of Chen \etal~\cite{chen2019learning} considers the exploration task and constructs the top-view occupancy map by unprojecting 3D points observed in the depth images. The  egocentric map is then passed to an exploration policy. 
With the exception of~\cite{gordon2018iqa}, the learned maps do not consider semantic and contextual information which has been shown to be important in learning generalizable navigation policies~\cite{mousavian2019visual,das2018embodied}. 

The effectiveness of semantic component, semantic segmentation and object detection has been in approaches which use recurrent neural networks
~\cite{fang2019scene, khan2017memory}. For example, Fang \etal~\cite{fang2019scene} uses a scene memory comprised of separately embedded observations at different time steps. While this scene memory encodes semantic information, the lack of structure neglects the spatial configurations of the objects and other semantic categories in the scene.

Our work is also related to large body of work on target driven navigation. The existing approaches differ in the level of supervision, model architectures and tasks.
~\cite{mousavian2019visual, sadeghi2019divis, ye2018active, zhu2017target, ye2019gaple, das2018embodied, mirowski2016learning,Wijmans_2019_CVPR, House3D_NIPS17}. For instance, Mousavian \etal~\cite{mousavian2019visual} and Ye \etal~\cite{ye2019gaple} learn effective representations for navigation using the outputs of object detectors and semantic segmentations in order to enable better generalization to novel environments and consider navigation policies, with state modelled by LSTM.  
Sadeghi \etal~\cite{sadeghi2019divis} focuses on collision avoidance for the goal reaching task and relies on convolutional LSTM to keep track of the goal's position with respect to the agent.
The works of Ye \etal~\cite{ye2018active} and Zhu \etal~\cite{zhu2017target} use deep reinforcement learning 
to train room-type specific navigation policies using feed-forward architectures, where the goal is provided as an image cropped from the scene. 
Finally, Das \etal~\cite{das2018embodied} uses embeddings of  a feed-forward model pre-trained on various tasks (i.e. semantic segmentation, depth prediction) as input observation for training
the policy. 
Even though these methods usually include semantic information as an input, they do not explicitely store in a spatial memory and use LSTM modules to retain the history of observations and actions. 



\begin{figure*}[!ht]
\begin{center}
\includegraphics[width=1\linewidth]{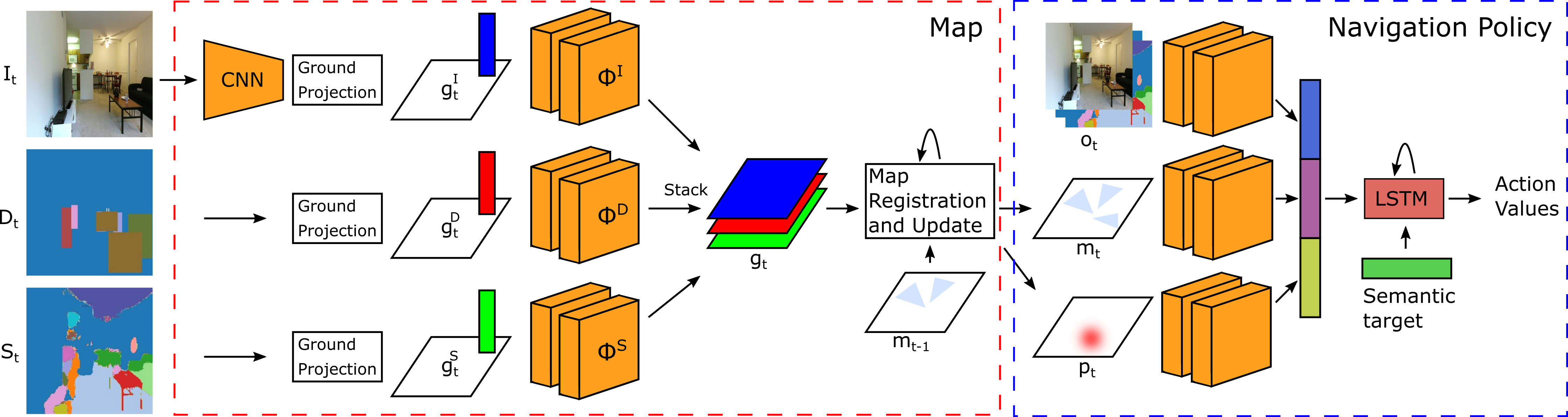}
\end{center}
   \caption{Overview of our simultaneous target driven navigation and mapping approach for a single timestep. We use as inputs the egocentric observations RGB image $I_t$, the detection masks $D_t$, and semantic segmentation $S_t$. Each input is first projected to a ground grid before extracting a feature embedding from each grid location. The grids are stacked and passed through a recurrent map registration and update module (see text for more details) which provides the updated map $m_t$ and localization prediction $p_t$. These, along with the egocentric observations $o_t$ are passed to a navigation module that extracts and concatenates their embeddings with the semantic target. Finally, the embeddings are passed to an LSTM that predicts the values for the next actions.
   Orange color signifies convolutional blocks, while other colors in the figure denote other feature representations.}
\vspace{-1.5em}
\label{fig:architecture}
\end{figure*}

\section{Approach}
Here we present the two modules of our method. First, we describe in detail the spatial map and how it is endowed with semantic information. Then, we outline the navigation policy and explain how the map is used in its training. An overview of the proposed architecture is in Figure~\ref{fig:architecture}.

\subsection{Learned Semantic Map}
We are interested in building an allocentric spatial map that encodes the agent's experiences during navigation episodes. Following MapNet~\cite{henriques2018mapnet} formulation, the map at time $t$ is represented as a grid $m_t \in \mathbb{R}^{u \times v \times n} $ of spatial dimensions $u \times v$ with feature embedding of size $n$ at each grid location. Besides an RGB image representation, we also extend the map to include the semantic information from object detection masks and semantic segmentation. The inputs are initially projected to an egocentric ground grid $g_t \in \mathbb{R}^{u' \times v' \times n}$. For the projection we use the available depth image and the camera intrinsic parameters to obtain a 3D point cloud from the image points. Each 3D point is then mapped to one of the $u' \times v'$ grid coordinates: $x_g = \lfloor \frac{x}{x_b} \rfloor + \frac{u'-1}{2}$, $z_g = \lfloor \frac{z}{z_b} \rfloor + \frac{v'-1}{2}$, where $x_g,z_g$ are the grid coordinates, $x_b,z_b$ are the dimensions of each bin in the grid, and $x,z$ are coordinates of the 3D point. The $y$ coordinate corresponding to the height of the point is neglected in this version of our work. Since multiple 3D points project to the same grid cell, the projected inputs are pooled to form a single vector. Specifically for each input type we get a grid as follows:\\
\noindent
\textbf{RGB image.} Given an input image $I_t$, we obtain a feature map $x_t \in \mathbb{R}^{h \times w \times n'}$ from any backbone CNN (e.g. VGG-16, ResNet50). In order to aggregate the features from different image regions we perform max-pooling over all features vectors projected to the same grid cell to yield the final grid $g_t^I \in \mathbb{R}^{u' \times v' \times n'}$. \\
\noindent
\textbf{Detection mask.} We run Faster R-CNN~\cite{fasterRCNN_2015} which is pre-trained on COCO~\cite{linCOCO2014} and convert the detections to $h \times w \times c_d$ binary bounding box masks. Each channel has detection masks of a particular class from COCO, where $c_d$ is the number of available classes. We get the grid $g_t^D \in \mathbb{R}^{u' \times v' \times c_d}$ by averaging over the occurrences of each detected class in a bin.\\
\textbf{Semantic segmentation.} We use the model of~\cite{Mousavian3DV16} trained on NYUv2 dataset~\cite{NYUV2} that outputs a $h \times w$ semantic segmentation of an image. Each pixel takes a value between 0 and $c_s-1$ where $c_s$ is the number of classes in the NYUv2 dataset. The grid  $g_t^S \in \mathbb{R}^{u' \times v' \times c_s}$ for this observation holds a probability distribution over the semantic labels in each bin. 
\noindent
Different inputs create separate grids, which are then passed through a small CNNs, comprised of two convolutional layers providing per grid cell feature embedding for each input. This step is deliberately applied on the grids rather than the images directly, such that the learned embeddings can capture spatial dependencies present in the map grid. We then stack the outputs of the small CNNs to form the egocentric 2D grid  $g_t$ at time $t$:
\begin{equation}
    g_t = \big[ \phi_I(g_t^I), \phi_D(g_t^D), \phi_S(g_t^S) \big]
\end{equation}
where $\phi_I$, $\phi_D$, and $\phi_S$ denote small CNNs applied to embeddings of RGB image, detection masks, and semantic segmentation. The details about choices of individual parameters are described in the experimental sections. 

Given this semantically informed representation, we follow the strategy of~\cite{henriques2018mapnet} for localization and registration stage. 
In order to register $g_t$ in the current map $m_{t-1}$ we densely match $g_t$ with $m_{t-1}$ over all possible locations $u \times v$ and over multiple orientations $r$. This operation is carried out through cross-correlation (equivalent to convolution in deep learning literature) and produces a tensor $p_t \in \mathbb{R}^{u \times v \times r}$ of scores which denotes the likelihood of the agent's position and orientation in the map at time $t$. 
In practice, multiple rotated copies of $g_t$ are stacked 
together to obtain a $g'_t \in \mathbb{R}^ {u' \times v' \times n \times r}$ tensor. 
After the cross-correlation, the output is passed through a softmax activation function to get $p_t$. Before inserting $g_t$ in the map, we need to rotate it and translate it according to its localization prediction $p_t$. This is achieved through a deconvolution operation between $g'_t$ and $p_t$ that can be seen as a linear combination of $g'_t$ weighted by $p_t$. The result is a tensor that contains the egocentric grid observations at time $t$, and is aligned to and has the same dimensions as $m_{t-1}$.

Finally, localized egocentric map $g_t$ is used to update the current map $m_{t-1}$ using a long short-term memory unit (LSTM). Each location's feature embedding is passed through LSTM and updated independently. We have also experimented with other update methods, such as averaging the features, but found the informed updating due to LSTM's trainable parameters to be superior. This can be also attributed to the fact that the LSTM learns how to combine the embeddings of different modalities that comprise $g_t$ in order to be more effective during localization. 
The model is trained using localization loss.\\
\noindent
\textbf{Localization loss.} We use cross-entropy loss to supervise the prediction of $p_t$:
\begin{equation}
    L_{loc} = - \frac{1}{T} \sum_t^T \sum_k^K \hat{p}_{tk} \log p_{tk}
\end{equation}
where $\hat{p}_{t}$ is a one-hot vector representing the ground-truth pose, $T$ is the length of an episode, and $K=u \times v \times r$ is the number of classes corresponding to the discreet spatial locations and orientations in the map. We assume $p_0$ to be at the center of the map facing to the right, and all subsequent ground-truth poses are relative to $p_0$.

\subsection{Navigation Policy}\label{sec:navigation}
Our task involves navigation to a semantic target within unknown environment. Therefore, it can be formulated as a partially observable Markov decision process (POMDP) $(S, A, O, P(s' \vert s,a), R(s,a))$, where the state space $S$ consists of the agent's pose, action space $A$ consists of a discrete set of actions, and observation space $O$ is comprised of the egocentric RGB images. The reward $R(s,a)=d(s,c)-d(s',c)$ is defined as the progress towards the semantic target $c$ when at state $s$ the action $a$ is executed that leads to state $s'$, where $d(.,.)$ is the number of steps required on the shortest path between a state and the semantic target. Finally, $P(s'\vert s,a)$ represents the transition probabilities. 

We are interested in learning a policy that can leverage the rich semantic and structured information in the map. To this end, the input to our policy is the allocentric spatial map $m_t$ which holds all past experiences of the agent during the episode. Since we do not assume perfect localization, the map is accompanied by the pose prediction $p_t$ in order to help the policy to pay attention to the relevant parts of the map. 

The learned policy $\pi (a \vert o_t, m_t, p_t ; c)$ outputs a distribution over the action space given both egocentric observations $o_t \in O$ and the map (that can be thought as a spatial memory).
Since our focus is to navigate in novel environments, the map is being build as the agent moves along and the policy uses as input the accumulated map up to time $t$. 
The semantic target $c$ is represented as a one-hot vector over the set of classes. Finally, a collision indicator represented as a single bit is concatenated to the rest of the inputs in order to encourage the policy to recover after a collision.

\paragraph{Training.} Following the work of Mousavian et al~\cite{mousavian2019visual}, we train our policy model to predict the cost of each action $a$ at a certain state $s$ and a given target $c$ using an L1 loss:
\begin{equation}
    L_{nav} = \frac{1}{T|A|} \sum_t^T \sum_{a \in A} \big|y(o_t, m_t, p_t, a;c) - \hat{y}(s, a;c) \big|
\end{equation}
where $\hat{y}(s, a;c)=-R(s,a)$ is the ground-truth cost and
$y(o_t, m_t, p_t, a;c)$ is the predicted cost. Given the definition of $R(s,a)$, $\hat{y}(s, a;c)$ can only take one of three values; $-1$ if the action takes the agent one step closer to target, $1$ if it takes the agent one step further from the target, or $0$ if the distance remains unchanged. The last case is possible since there can be multiple target poses in an episode. If an action leads to a collision, then we assign $\hat{y}(s, a;c)=1$ even though the agent has not moved, while if an action leads to a goal we assign $\hat{y}(s, a;c)=-2$. 

The policy model is trained in a supervised fashion using an online variant of DAgger~\cite{dagger11}. In particular, we first generate training episodes by sampling a random starting point and target in a scene and selecting the actions along the shortest path (expert policy). At this stage we also sample trajectories along randomly chosen paths in order to increase the coverage of the observation space $O$ in the scenes. During training we sample the next minibatch either from the initially generated episodes (expert and random), or by unrolling the current policy to select new episodes. We start with a high probability of selecting from the initial episodes and gradually decrease this probability with exponential decay.

To accommodate this training paradigm, the environment is represented by a graph, where the nodes are discrete poses of the agent and edges represent possible actions. Each pose has corresponding RGB and depth images and the absence of edges between two nodes is treated as collision. In this setting, the shortest path between two nodes can be easily computed and used as supervision. 

\paragraph{Model architecture.} The policy $\pi (a \vert o_t, m_t, p_t ; c)$ is modeled by a convolutional NN. The image observations are first passed through a separate network that computes a 128 dimensional embedding.  The map $m_t$ and pose estimation $p_t$ are passed through a convolutional layer of $3 \times 3 \times 8$ followed by batch normalization and max-pooling of kernel size 2 and stride 2. The outputs are flattened and passed though a fully connected layer followed by dropout to get the embedding. 
For the egocentric observation $o_t$ we stack any available images (i.e. RGB, detection masks, or semantic segmentation) and use a pretrained ResNet18 (without the last layer) to extract 512 dimensional features, prior to computing the embedding. The one-hot vector that denotes the semantic target is also encoded to a 128 dimensional embedding.  
Then, all embeddings are concatenated and used as input to an LSTM layer of 512 units. Finally, a fully connected layer predicts the cost of each action $y(o_t, m_t, p_t, a;c)$.

\paragraph{Controller.} During inference, instead of directly choosing the action with the lowest cost, we sample from the predicted probability distribution over the actions by applying soft-max on the negated predicted costs. 
This still assigns the highest probability to the action with the predicted lowest cost, however it allows for some flexibility during decision making to avoid getting stuck in limited space situations.

\begin{table}
\begin{center}
\begin{tabular}{|c|c|c|c|}
\hline
Dataset & \multicolumn{2}{c|}{AVD} & \multicolumn{1}{c|}{Matterport3D} \\
\hline
Map Model & APE-5 & APE-20 & APE-20  \\
\hline
RGB & 285 & 800 & 993 \\
\hline
RGB-SSeg-Det & 215 & 692 & 803 \\ 
\hline
SSeg-Det & \textbf{179} & \textbf{647} & \textbf{655} \\
\hline
\end{tabular}
\end{center}
\caption{Localization results on the AVD and Matterport3D dataset using map models trained with different combinations of input modalities. The Average Position Error (APE) is reported in millimeters for episodes of length 5 and 20.}  
\label{tab:localization_res_avd}
\end{table}


\begin{figure}[!t]
\begin{center}
\includegraphics[width=0.32\linewidth]{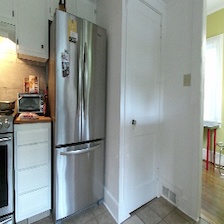}
\includegraphics[width=0.32\linewidth]{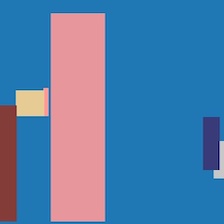}
\includegraphics[width=0.32\linewidth]{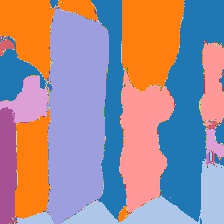}
\\
\includegraphics[width=0.32\linewidth]{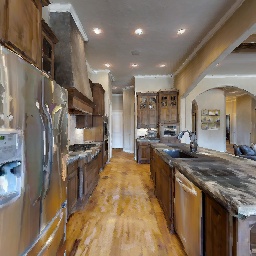}
\includegraphics[width=0.32\linewidth]{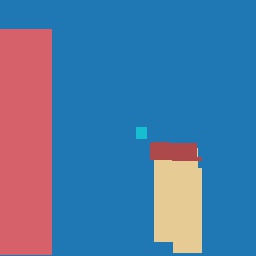}
\includegraphics[width=0.32\linewidth]{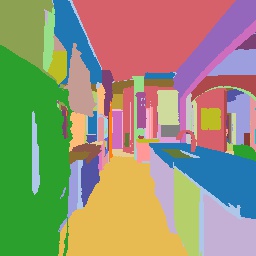}
\end{center}
   \caption{Example inputs from the AVD (top row) and Matterport3D (bottom row) datasets. From left to right we show the RGB image, detection masks and semantic segmentations.}
\label{fig:datasets}
\end{figure}

\begin{figure}[!t]
\begin{center}
\includegraphics[width=0.47\linewidth]{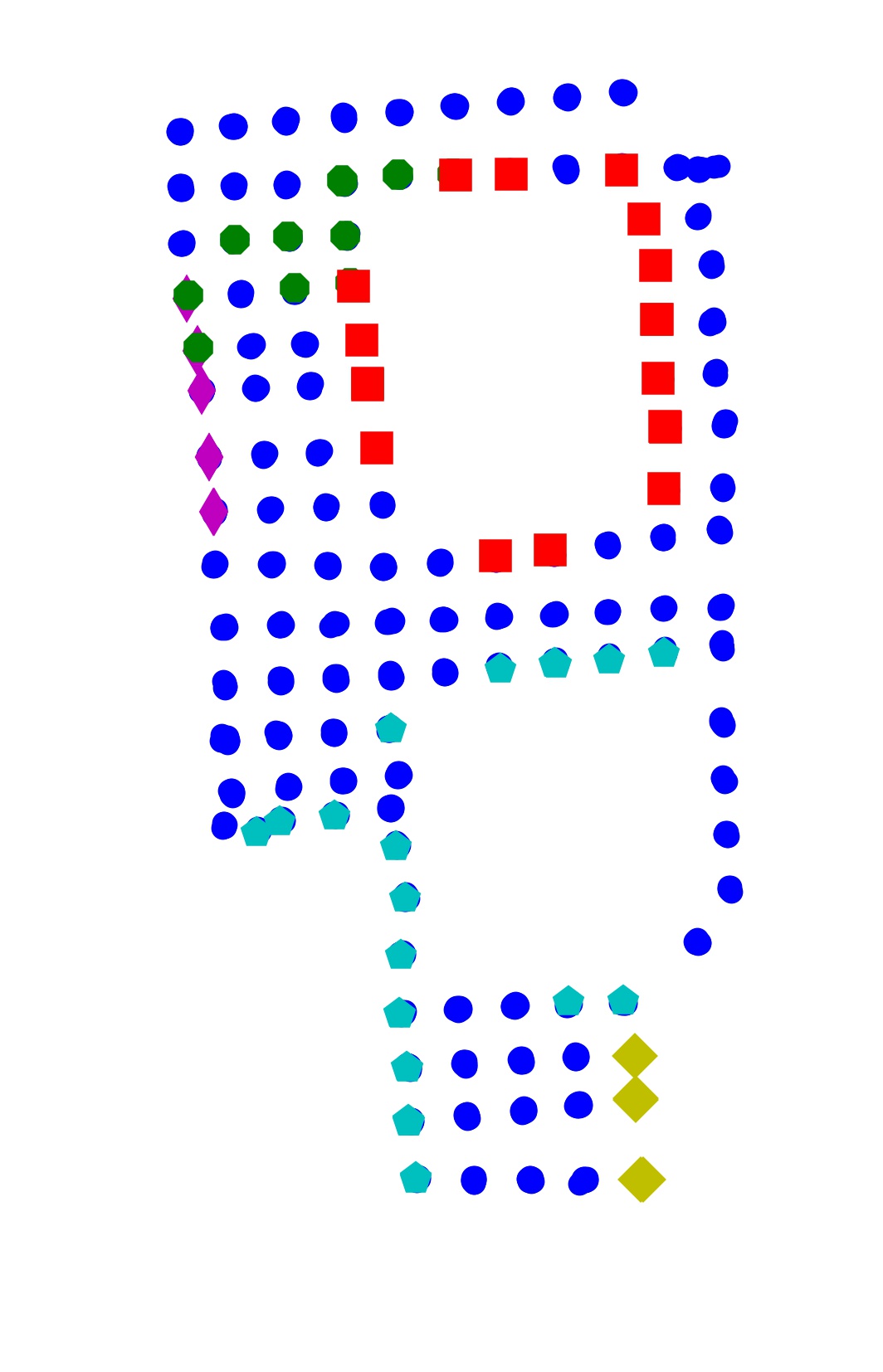}
\includegraphics[width=0.47\linewidth]{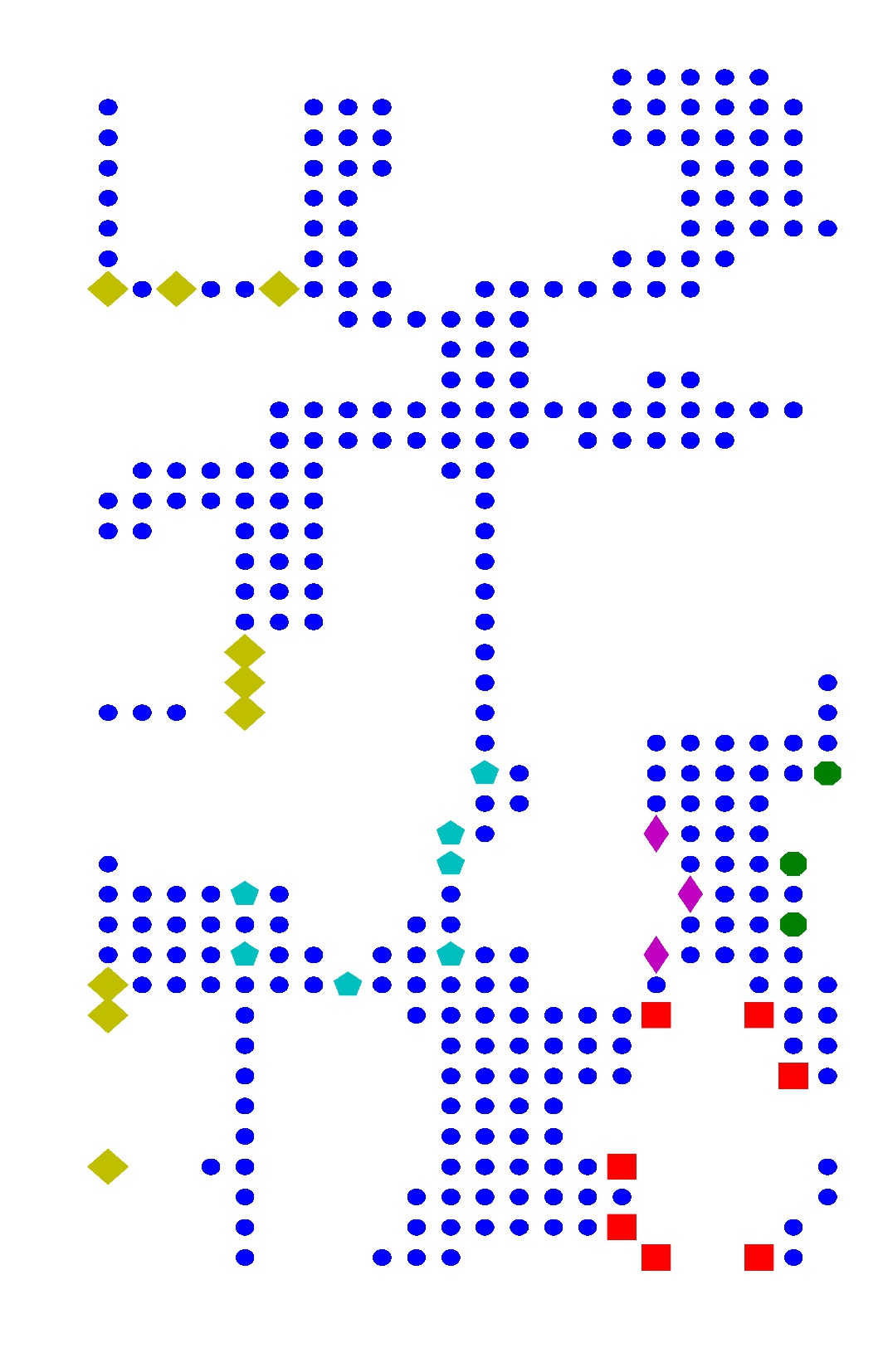}
\end{center}
   \caption{Visualizations of graphs along with target locations from an AVD scene (left) and a Matterport3D scene (right). The different shapes and colors denote different target objects.}
\label{fig:dataset_graphs}
\end{figure}

\section{Experiments}
We perform two main experiments, evaluation of the localization accuracy in the trained spatial map (sec. \ref{subsec:localization_exp}) and evaluation of the learned navigation policy on unknown environments (sec. \ref{subsec:navigation_exp}). The proposed method is demonstrated on two publicly available datasets, Active vision dataset (AVD)~\cite{active-vision-dataset2017} and Matterport3D~\cite{Matterport3D}. We illustrate input examples from the two datasets in Figure~\ref{fig:datasets}.


\textbf{AVD.} This dataset contains around 20,000 real RGB-D images from typical indoor scenes densely captured by a robot on a 2D grid every $30cm$. Each location on the grid offers multiple views at $30^\circ$ intervals. The data for each scene are organized as a graph where the edges are defined over a discrete set of actions. This provides with the ability to simulate the movement of an agent in a scene but with the luxury of having real images as observations.

\textbf{Matterport3D.} This dataset contains visually realistic reconstructions of indoor scenes with varying appearance and layout. We endow the dataset with the same structure as AVD by densely sampling navigable positions at $30cm$ intervals on the occupancy map of each scene. At each navigable position, we render RGB images, semantic segmentations and depth images from $12$ different orientations at $30^\circ$ intervals through Habitat-Sim~\cite{savva2019habitat}. The images are connected through actions based on their spatial neighborhood and orientations. The built dataset contains more than $10,000$ images for each scene, which is considerably larger than AVD. We use 17 scenes for training and 5 for testing. 


\subsection{Localization}\label{subsec:localization_exp}
To validate the effectiveness of using the semantic information for mapping, we present localization results for allocentric maps that were trained with different input combinations and episode lengths of 5 and 20 steps.
For both datasets an agent is simulated through random walks in order to collect $33,000$ and $88,000$ training episodes for AVD and Matterport3D respectively. The trained models are applied on episodes not seen during training and are evaluated on Average Position Error (APE), which measures the average Euclidean distance between the ground-truth pose and the predicted pose.

\paragraph{Implementation details.}  RGB image is passed through a pretrained truncated ResNet50 using only the first 11 layers. The small CNNs for each input grid modality ($\phi_I$, $\phi_D$, $\phi_S$) are realized with two convolutional layers of $3 \times 3 \times 64$ and $3 \times 3 \times l$, where $l=32$ for $\phi_I$, and $l=16$ for both $\phi_D$ and $\phi_S$. 
The number of units $n$ for the LSTM corresponds to the summation of the embedding dimensions of the modalities used for the particular experiment. 
Regarding the hyper-parameters of the map, we define the map dimensions $u=v=29$, the egocentric observation grid dimensions $u'=v'=21$, the number of rotations $r=12$, and the grid cell size $x_b=z_b=300mm$. 

\paragraph{Results and discussion.} The results are illustrated in Table~\ref{tab:localization_res_avd}, where  semantic segmentation is denoted as \textit{SSeg} and detection masks as \textit{Det}. There is no direct comparison to the results from~\cite{henriques2018mapnet} since the exact train and test sets are not provided, however our map model trained only with RGB can be considered analogous to the MapNet trained in ~\cite{henriques2018mapnet}. We observe that the model with the lowest localization error in both 5 and 20 step cases is  \textit{SSeg-Det}, which is not using RGB information. 
This can be attributed to the fact that the RGB image representation needs to capture view-invariant properties, which is difficult to achieve, such that the egocentric ground grid can be accurately matched to the allocentric map.
On the other hand, this is not necessary in the case of \textit{SSeg-Det}, since it operates on recognition outputs.
This is also highlighted on the Matterport3D results, where the images are synthetically generated and the average position error difference is more in favour of \textit{SSeg-Det}. 
This effectively demonstrates that a scene can be memorized with respect to only semantic information such that it is useful for re-localization.
In fact, when the RGB images are added then the error slightly increases. It is also important to note that the authors of~\cite{henriques2018mapnet} demonstrated superior performance of their approach with respect to traditional ORB-SLAM\cite{mur2015orb} on AVD using only RGB images. We exhibit that additional semantic information further improves the localization ability of the agent.  

\begin{table*}
\begin{center}
\begin{tabular}{|c|c|c|c|c|}
\hline
Dataset & \multicolumn{2}{c|}{AVD} & \multicolumn{2}{c|}{Matterport3D} \\
\hline
Method & Success Rate (\%) & Path Len. Rat. & Success Rate (\%) & Path Len. Rat. \\
\hline
Random Walk & 24.3 & 4.4 & 13.6 & 10.7 \\
\hline
Non-learning & 35.3 & 9.3 & 42.8 & 3.6 \\
\hline
No-Map-AVD~\cite{mousavian2019visual} & 48.0 & - & - & - \\
\hline 
No-Map-SUNCG~\cite{mousavian2019visual} & 54.0 & - & - & - \\ 
\hline
Ours & \textbf{64.6} & \textbf{1.9} & \textbf{69.5} & 2.3 \\
\hline
\end{tabular}
\end{center}
\caption{Results of semantic target navigation in novel scenes in AVD and Matterport3D. }  
\label{tab:navigation_res_1}
\end{table*}



\begin{figure*}[!t]
\begin{center}
\includegraphics[width=1\linewidth]{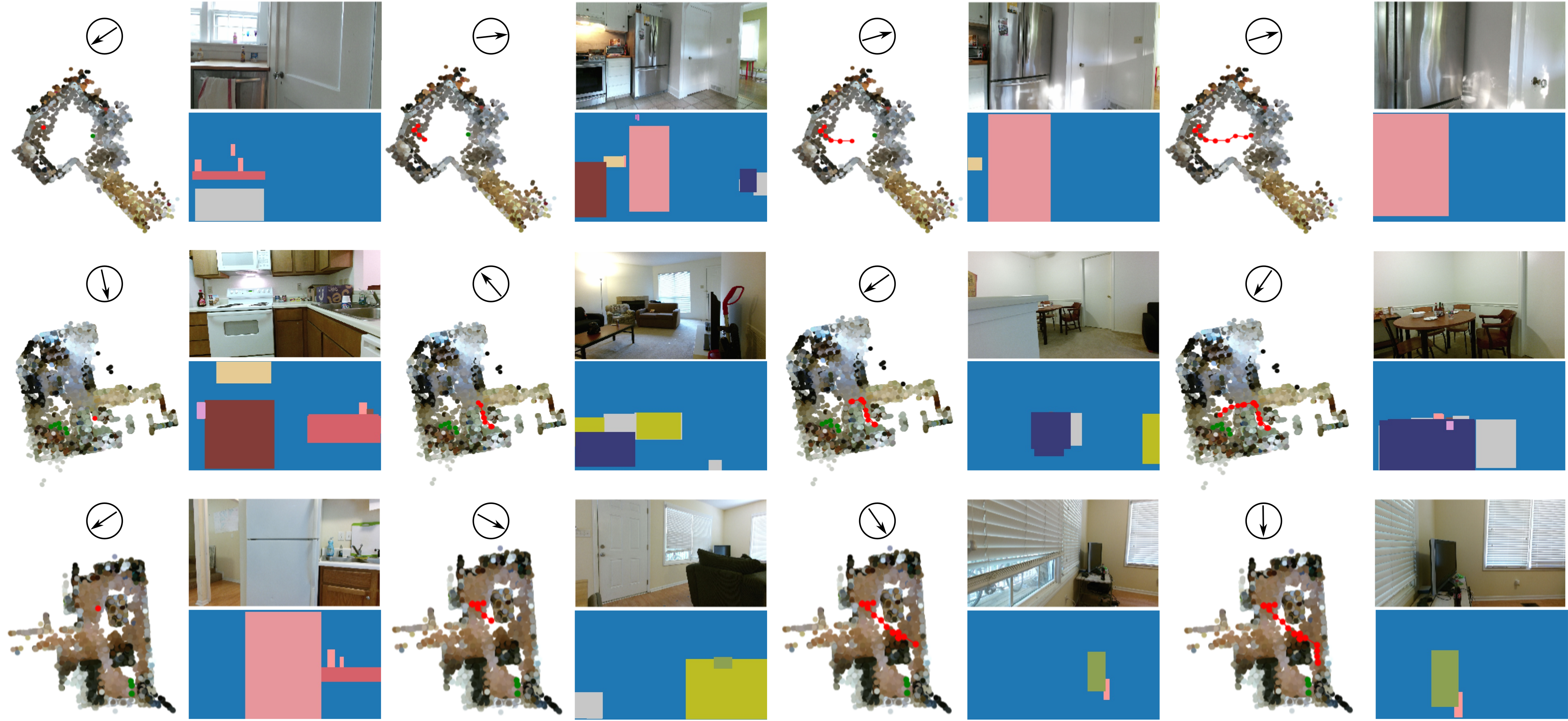}
\end{center}
   \caption{Qualitative navigation results on the AVD dataset. Each row corresponds to a different episode. From top to bottom, the target object is the fridge, dining table, and TV. For each step of an episode we present the map with the agent's trajectory up to that time, the agent's orientation, RGB image and detection masks. Notice that in all three episodes the agent moves quickly towards the target once it is detected and placed in the map.}
\label{fig:navigation_AVD}
\end{figure*}

\subsection{Navigation to semantic target}\label{subsec:navigation_exp}
Here we investigate the effectiveness of the proposed target driven navigation policy. Our objective in this experiment is twofold. First, we would like to demonstrate the effect of using a spatial map in comparison to LSTM policies that do not use a map, and second, investigate navigation policies that are learned with spatial maps of different modalities. 

The training procedure is as follows. First, the spatial map model is trained with the localization objective. The training of the navigation policy uses the frozen mapping module to update the map and predict the agent's pose at each step. During this procedure, the mapping module is fine-tuned through the navigation objective. Note that at the beginning of each episode the map contains no information. Unless otherwise specified, we train the navigation policy using the \textit{SSeg-Det} map model for AVD, while the \textit{RGB-SSeg-Det} map model is utilized when learning the policy on Matterport3D.

For both AVD and Matterport3D five semantic targets are identified: \textit{\{dining\_table, refrigerator, tv, couch, microwave\}}. Figure~\ref{fig:dataset_graphs} presents examples of scene graphs with marked target objects.
In the case of AVD we compare our approach to~\cite{mousavian2019visual}, therefore we follow their train/test split of \textit{different environments} (11 scenes for training and 3 for testing), and use the target locations of object categories they provide. As mentioned in sec.~\ref{sec:navigation} the training data are generated using DAGGER with 55,000 and 88,000 initial training episodes for AVD and Matterport3D respectively. 
For evaluation, the percentage of successful episodes is reported along with the average path length ratio. The episode is successful when the agent is at least 5 steps away from a target pose. The path length ratio is the ratio between the predicted path and the shortest path length and is calculated only for successful episodes. The maximum number of steps for each episode is 100.

\paragraph{Comparison to other policies.}
To demonstrate our method's superiority to policies which do not use a spatial map we have defined three baselines:\\
\noindent
\textit{Random walk.} The agent chooses random actions until it reaches the target.\\
\noindent
\textit{Non-learning baseline.}  Similarly to~\cite{mousavian2019visual}, the agent chooses random actions until the target object is detected, in which case the agent computes the shortest path to the target. Note that for this baseline, the agent has full knowledge of the environment and its graph once it detects the target.\\
\noindent
\textit{Learned LSTM policy without a map.} The method proposed in~\cite{mousavian2019visual}. The policy is learned using only egocentric observations without any spatial memory. We compare to a policy trained on AVD that uses detection masks denoted as \textit{No-Map-AVD} and policy trained on both AVD and the large synthetic dataset SUNCG~\cite{song2017semantic} that uses detection masks and semantic segmentations, here referred to as \textit{No-Map-SUNCG}. This is the best performing method reported in~\cite{mousavian2019visual} as it leverages the vast amounts of data offered in SUNCG's 200 synthetic environments.

Results are shown in Table~\ref{tab:navigation_res_1}. On AVD dataset the presented approach outperforms the second best method by $10.6\%$, without any use of complementary synthetic data during training. 
The biggest advantage of our method compared to the other learned baselines is that our agent has access to semantic information stored in a structured memory that corresponds to the history of observations. This reduces the amount of information that LSTM retains and helps during the optimization of the policy.
Furthermore, the lower average path length ratio than the \textit{Non-learning} baseline, in both datasets, suggests that our navigation model learns contextual cues from the semantic map that reduce the time spent searching for the target. In addition, as demonstrated in Figure~\ref{fig:navigation_AVD}, the agent learns to quickly move towards the target upon detection. Note that the baseline uses an optimal path when the target is detected.
Note that the average path length ratio result is not directly comparable to the other learned baselines since they require a \textit{stop} action to end an episode, which we do not use. 


\begin{table}
\begin{center}
\begin{tabular}{|l|c|}
\hline
Model & Success Rate (\%) \\
\hline
\hline
1. Nav-RGB & 60.1 \\
\hline 
2. Nav-RGB-Det & 61.1  \\ 
\hline
3. Nav-RGB-SSeg-Det & 63.2  \\
\hline
4. Nav-SSeg-Det & \textbf{64.6}  \\
\hline
\hline
5. Nav-RGB-NF & 58.2 \\
\hline
6. Nav-SSeg-Det-NF & 60.4 \\ 
\hline
\hline
7. Nav-RGB-NF-NE & 53.9 \\
\hline
8. Nav-SSeg-Det-NF-NE & 56.9  \\ 
\hline
\end{tabular}
\end{center}
\caption{Results of our ablation study on AVD, illustrating the performance of navigation models trained with different map models, without fine-tuning the map (NF), or without using any egocentric observations (NE).}  
\label{tab:ablation_res_1}
\end{table}

\begin{table}
\begin{center}
\begin{tabular}{|l|c|}
\hline
Model & Success Rate (\%) \\
\hline
\hline
1. Nav-SSeg-Det & 62.9  \\
\hline
2. Nav-RGB & 66.7 \\
\hline
3. Nav-RGB-SSeg-Det & \textbf{69.5} \\
\hline
\hline
4. Nav-SSeg-Det-NF & 59.7 \\ 
\hline
5. Nav-RGB-NF & 64.2 \\
\hline
\end{tabular}
\end{center}
\caption{Results of our ablation study on Matterport3D, illustrating the performance of navigation models trained with different map models and without fine-tuning the map (NF).}  
\label{tab:ablation_res_2}
\end{table}

\paragraph{Ablation Study}
In this section we attempt to get a better understanding of how certain components of our method contribute to the overall performance.

\textit{Variations of spatial map modalities.} Here we pose the question of which are the most suitable map feature representations when learning to navigate. To this end, we trained multiple navigation models using spatial maps that were learned with different modality inputs. 
Results are presented in lines 1-4 of Table~\ref{tab:ablation_res_1} for AVD, and lines 1-3 of Table~\ref{tab:ablation_res_2} for Matterport3D. In the case of AVD, we notice that the \textit{Nav-SSeg-Det} outperforms \textit{Nav-RGB} by $4.5\%$. This validates our assumption that navigating to a semantic target can be successful using a map with purely semantic features. Note also that the map model \textit{SSeg-Det} demonstrated the lowest localization error (see Table~\ref{tab:localization_res_avd}). However, for Matterport3D we observe that \textit{Nav-RGB} outperforms \textit{Nav-SSeg-Det}, while the best model is the one using all modalities (\textit{Nav-RGB-SSeg-Det}).
This can be explained by the fact that the detections are very noisy due to the artifacts in Matterport3D's images. Hence, they are not as reliable when training the policy. Another reason could be that Matterport3D has larger scenes and object encounters are sparser, therefore providing less useful information in the map.  

\textit{Joint training.} To see the effect of fine-tuning the mapping module we re-trained selected navigation policies but kept the map network parameters frozen. The results are in lines 5, 6 of Table~\ref{tab:ablation_res_1} for AVD and lines 4, 5 of Table~\ref{tab:ablation_res_2} for Matterport3D, where the models trained without fine-tuning are denoted as $NF$. There is a consistent reduction of performance when fine-tuning is not performed in both datasets. This shows that the map embeddings learned during the initial training of the map module are not immediately applicable for navigation and further adjustment is required.

\textit{Effect of egocentric observation.} We rely mainly on the allocentric map and pose prediction to decide future actions, which requires a depth sensor during the ground projection step. However, in cases where the agent is very close to an object and therefore outside of the effective range of the depth sensor, the projection is unreliable. We argue that using complementary egocentric observations as input to the navigation policy can help mitigate this problem. Results of models trained with and without egocentric observations are reported in lines 7, 8 of Table~\ref{tab:ablation_res_1} for AVD. We observe that there is a rough $4\%$ degradation in the performance compared to lines 5, 6 in the same table, which validates our assumption.


\section{Conclusions}
We have presented a new method for simultaneous mapping and target driven navigation in novel environments. The mapping component leverages the outputs of object detection and semantic segmentation to construct a spatial representation of a scene  which contains some semantic information. 
This representation is then used to optimize a navigation policy that takes advantage of the agent's experiences during an episode encoded in the allocentric spatial map. 
The experiments on AVD and Matterport3D environments demonstrate that our approach outperforms only RGB baselines for the task of localization, and non-mapping baselines for the target-driven navigation.

{\small
\bibliographystyle{ieee_fullname}
\bibliography{egbib}
}

\end{document}